\documentclass[preprints,article,accept,moreauthors,pdftex,10pt,a4paper]{Definitions/mdpi}

\usepackage{subfigure}

\makeatletter
\newif\if@restonecol
\makeatother

\usepackage[linesnumbered,ruled,vlined]{algorithm2e}%[ruled,vlined]{
\usepackage{algpseudocode}
\usepackage{amsmath}
\usepackage{bm}
\usepackage{mathrsfs}
\usepackage{amsfonts,amssymb}
\RequirePackage[normalem]{ulem} %DIF PREAMBLE
\RequirePackage{color}\definecolor{RED}{rgb}{1,0,0}\definecolor{BLUE}{rgb}{0,0,1} %DIF PREAMBLE
\providecommand{\DIFadd}[1]{{\protect\color{black}{#1}}} %DIF PREAMBLE
\providecommand{\DIFdel}[1]{{\protect\color{red}\sout{}}}                      %DIF PREAMBLE
\providecommand{\DIFaddbegin}{} %DIF PREAMBLE
\providecommand{\DIFaddend}{} %DIF PREAMBLE
\providecommand{\DIFdelbegin}{} %DIF PREAMBLE
\providecommand{\DIFdelend}{} %DIF PREAMBLE
\providecommand{\DIFaddFL}[1]{\DIFadd{#1}} %DIF PREAMBLE
\providecommand{\DIFdelFL}[1]{\DIFdel{#1}} %DIF PREAMBLE
\providecommand{\DIFaddbeginFL}{} %DIF PREAMBLE
\providecommand{\DIFaddendFL}{} %DIF PREAMBLE
\providecommand{\DIFdelbeginFL}{} %DIF PREAMBLE
\providecommand{\DIFdelendFL}{} %DIF PREAMBLE
\firstpage{1}
\pubvolume{xx}
\issuenum{0}
\articlenumber{00}
\pubyear{2019}
\copyrightyear{2019}
\history{Received: date; Accepted: date; Published: date}
\Title{A Novel Self-Intersection Penalty Term for Statistical Body Shape Models and Its Applications in 3D Pose~Estimation}

\Author{Zaiqiang Wu $^{1}$, Wei Jiang $^{1,}$*, Hao Luo $^{1}$ and Lin Cheng $^{2}$}
\AuthorNames{Zaiqiang Wu, Wei Jiang, Hao Luo and Lin Cheng}

\address{%
$^{1}$ \quad College of Control Science and Engineering, Zhejiang University, Hangzhou 310007, China; wuzaiqiang@zju.edu.cn (Z.W.); haoluocsc@zju.edu.cn (H.L.)\\
$^{2}$ \quad 2012 Lab, Huawei Technologies, Hangzhou 310028, China; chenglin17@huawei.com}
\corres{Correspondence: jiangwei\_zju@zju.edu.cn; Tel.: +86-150-6818-8247}

\abstract{Statistical body shape models are widely used in 3D pose estimation due to their low-dimensional parameters representation. However, it is difficult to avoid self-intersection between body parts accurately. Motivated by this fact, we proposed a novel self-intersection penalty term for statistical body shape models applied in 3D pose estimation. To avoid the trouble of computing self-intersection for complex surfaces like the body meshes, the gradient of our proposed self-intersection penalty term is manually derived from the perspective of geometry. First, the self-intersection penalty term is defined as the volume of the self-intersection region. To calculate the partial derivatives with respect to the coordinates of the vertices, we employed detection rays to divide vertices of statistical body shape models into different groups depending on whether the vertex is in the region of self-intersection. Second, the partial derivatives could be easily derived by the normal vectors of neighboring triangles of the vertices. Finally, this penalty term could be applied in gradient-based optimization algorithms to remove the self-intersection of triangular meshes without using any approximation. Qualitative and quantitative evaluations were conducted to demonstrate the effectiveness and generality of our proposed method compared with previous approaches. The~experimental results show that our proposed penalty term can avoid self-intersection to exclude unreasonable predictions and improves the accuracy of 3D pose estimation indirectly. Further more, the proposed method could be employed universally in triangular mesh based 3D reconstruction.} %DIF >
 %DIF >
\begin{document}

%%%%%%%%%%%%%%%%%%%%%%%%%%%%%%%%%%%%%%%%%%
%% Only for the journal Gels: Please place the Experimental Section after the Conclusions

%%%%%%%%%%%%%%%%%%%%%%%%%%%%%%%%%%%%%%%%%%
%\setcounter{section}{-1} %% Remove this when starting to work on the template.
%\section{How to Use this Template}
%The template details the sections that can be used in a manuscript. Note that the order and names of article sections may differ from the requirements of the journal (e.g., the positioning of the Materials and Methods section). Please check the instructions for authors page of the journal to verify the correct order and names. For any questions, please contact the editorial office of the journal or support@mdpi.com. For LaTeX related questions please contact Janine Daum at latex-support@mdpi.com.
%The order of the section titles is: Introduction, Materials and Methods, Results, Discussion, Conclusions for these journals: aerospace,algorithms,antibodies,antioxidants,atmosphere,axioms,biomedicines,carbon,crystals,designs,diagnostics,environments,fermentation,fluids,forests,fractalfract,informatics,information,inventions,jfmk,jrfm,lubricants,neonatalscreening,neuroglia,particles,pharmaceutics,polymers,processes,technologies,viruses,vision

\section{Introduction}
Estimating a 3D human pose from a single 2D image, and more generally, reconstructing the 3D model from 2D images is one of the fundamental and challenging problems in 3D computer vision due to the inherent ambiguity in inferring 3D from 2D. Choosing the appropriate 3D representation is vital for 3D reconstruction. There are many types of 3D representations for 3D modeling. Voxels, point clouds and polygon meshes are commonly used 3D formats for 3D representation. Voxels can be fed directly to convolutional neural networks (CNNs), therefore a lot of works applied voxels for classification \cite{maturana2015voxnet, qi2016volumetric} and 3D reconstruction \cite{choy20163d, tatarchenko2017octree, tulsiani2017multi}. However voxels are poor in memory efficiency. To avoid this drawback of voxel representation, Fan et al. \cite{fan2017point} proposed a method to generate point clouds from 2D images. But since there are no connections between points in the point cloud representation, the generated point cloud is often not close to a surface. Polygon mesh is promising due to its  high memory efficiency when compared to voxels and point clouds \cite{kato2018neural}. Polygon mesh is also convenient to visualize since it is compatible with most existing rendering engines.

There are many works using polygon meshes, especially triangular meshes, to represent 3D pose estimation results. Anguelov et al. \cite{anguelov2005scape} proposed the first \DIFdelbegin \DIFdel{statical }\DIFdelend \DIFaddbegin \DIFadd{statistical }\DIFaddend body shape model called SCAPE, %Please define if appropriate.
 represented as a triangular mesh. Loper et al. \cite{loper2015smpl} proposed another \DIFdelbegin \DIFdel{statical }\DIFdelend \DIFaddbegin \DIFadd{statistical }\DIFaddend body shape model with higher accuracy called SMPL %Please define if appropriate.
  which is also represented as a triangular mesh. \mbox{Guan et al.}~\cite{guan2009estimating, guan2012virtual} employed SCAPE to estimate 3D poses based on manually marked 2D joints. \mbox{Bogo et al.}~\cite{bogo2016keep} employed the SMPL human model and minimized the error between the projected human model joints and 2D joints detected by DeepCut \cite{pishchulin2016deepcut} to estimate the 3D pose of the human body automatically, this method is iterative optimization-based which results in high accuracy but is time consuming. \mbox{Pavlakos et al.}~\cite{pavlakos2018learning} used a variant of Hourglass \cite{newell2016stacked} to predict 2D joints and 2D masks simultaneously, then the 2D joints and 2D masks were used to regress pose parameters and shape parameters of SMPL separately in a direct prediction way. This approach is much faster than method proposed in \cite{bogo2016keep}, however self-intersection occurs on images with pattern of poses that never appeared in the training~set.

Other work utilized triangle meshes to represent 3D reconstruction results of objects. \mbox{Kar et al.}~\cite{Kar_2015_CVPR} trained a mesh deformable model to reconstruct 3D shapes limited to the popular categories. \mbox{Kato et al.}~\cite{kato2018neural} deformed a predefined mesh to approximate the 3D object by minimizing the silhouette error. Wang et al. \cite{wang2018pixel2mesh} proposed an end-to-end deep learning architecture which represents a triangular mesh in a graph-based convolutional neural network to estimate the 3D shape of objects from a single~image.

\begin{figure}[H]
\centering
\DIFdelbeginFL %DIFDELCMD < \includegraphics[width=16 cm,trim={0 8cm 0 6cm},clip]{figures/example}
%DIFDELCMD < %%%
\DIFdelendFL \DIFaddbeginFL \includegraphics[width=16 cm,trim={1.2cm 7cm 0 8cm},clip]{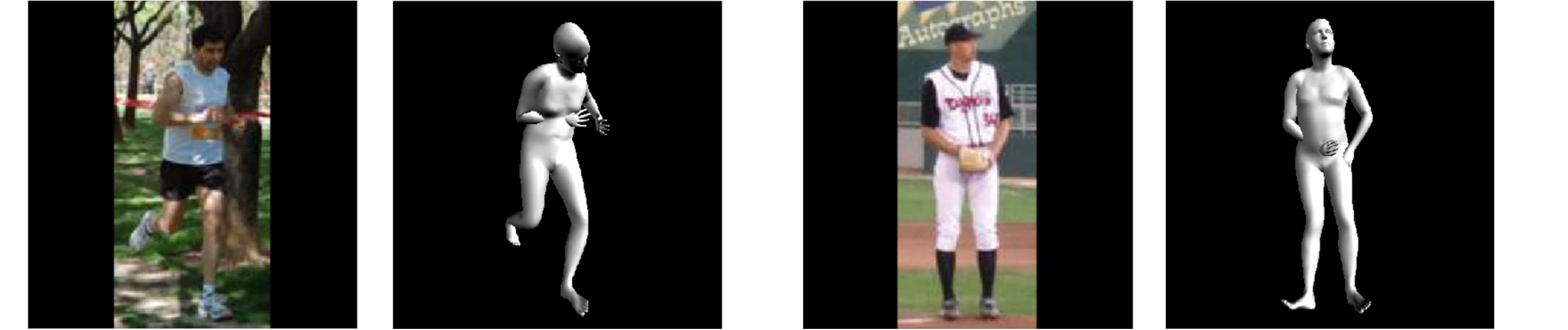}
\DIFaddendFL \caption{Examples of model-based 3D pose estimation results with self-intersection between body~parts.}
\label{fig:example}
\end{figure}

However, one of the main disadvantages of representing 3D shapes as triangular meshes is that self-intersection is difficult to prevent. Some examples of 3D pose estimation results with self-intersection are shown in Figure \ref{fig:example}. It is impossible for objects in the real world to have surfaces with self-intersection. Therefore previous work paid great attention to avoiding the self-intersection of triangular meshes. Since it is difficult to derive a differentiable expression of the intersection volume directly, \DIFdelbegin \DIFdel{so }\DIFdelend approximation is often taken \DIFaddbegin \DIFadd{to simplify the derivation}\DIFaddend . Although approximation \DIFdelbegin \DIFdel{simplifies the problem of }\DIFdelend \DIFaddbegin \DIFadd{provides great convenience to }\DIFaddend deriving a differentiable penalty term, self-intersection can not be removed strictly since approximation can not describe the original surface accurately.

To overcome the weakness of previous methods of preventing self-intersection, this paper proposed a novel self-intersection penalty term (SPT) which \DIFaddbegin \DIFadd{is able to }\DIFaddend avoid self-intersection strictly. Unlike \DIFdelbegin \DIFdel{traditional approaches, the }\DIFdelend \DIFaddbegin \DIFadd{previous approaches, our proposed }\DIFaddend self-intersection penalty term is defined as the volume of intersection regions which is expensive to compute \DIFaddbegin \DIFadd{however, we managed to work around this problem}\DIFaddend . Notably, no approximation was taken in this paper to derive a differentiable expression. Besides, it is not worthwhile to derive a differentiable expression of intersection volume since calculating the exact volume is not our intention. Moreover, the partial derivatives can be easily derived even without the expression of intersection volume. Inspired by \cite{crow1977shadow}, we developed an algorithm to detect vertices of self-intersection regions quickly by only going through triangles in the mesh once. This process is similar to rasterization in computer graphic with linear time complexity. A linked list is applied to store depth values and \DIFdelbegin \DIFdel{orientation }\DIFdelend \DIFaddbegin \DIFadd{orientations }\DIFaddend of triangles intersected with the same detection ray. Then vertices in self-intersection region can be easily detected by going through the linked list. The partial derivatives of the self-intersection term with respect to the coordinate of each vertex \DIFdelbegin \DIFdel{is }\DIFdelend \DIFaddbegin \DIFadd{are }\DIFaddend easy to derive from the perspective of geometry. The partial derivatives with respect to vertices not in the region of self-intersection is obviously zero, while the partial derivatives with respect to vertices in the region of self-intersection could be obtained from the normal vectors of neighbouring triangles. The value of the penalty term is assigned as the ratio of number of vertices in self-intersection to number of vertices not in self-intersection. In this way the value of penalty term is easy to compute and indicates the degree of self-intersection in some degree. The experimental results show that the proposed penalty term avoids self-intersection strictly and works effectively. The main contributions of this paper are summarized as follows:
\begin{itemize}[leftmargin=*,labelsep=5.8mm]
\item	We proposed a novel self-intersection penalty term which does not require deriving a differentiable expression and generally applies to triangular mesh-based representations.
\item   We performed 3D pose estimation from 2D joints \DIFaddbegin \DIFadd{and compared our method with other state-of-the-art approaches qualitatively and quantitatively }\DIFaddend to demonstrate the practical value \DIFaddbegin \DIFadd{and the superiority }\DIFaddend of our method.
\item	To the best of our knowledge, this is the first time the conception of self-intersection in relation to disconnected meshes in the field of 3D reconstruction has been generalized%Please confirm meaning has been retained.
.
\end{itemize}

The content of this paper consists of five sections. In Section \ref{sec:relatedwork} an overview of related work is provided. In Section \ref{sec:method}, the details of the proposed self-intersection penalty term are presented. In~Section~\ref{sec:experiment}, the results of experiments and analysis of the proposed self-intersection penalty term are given. The conclusions are presented in Section \ref{sec:conclusion}.
%The introduction should briefly place the study in a broad context and highlight why it is important. It should define the purpose of the work and its significance. The current state of the research field should be reviewed carefully and key publications cited. Please highlight controversial and diverging hypotheses when necessary. Finally, briefly mention the main aim of the work and highlight the principal conclusions. As far as possible, please keep the introduction comprehensible to scientists outside your particular field of research. Citing a journal paper \cite{bogo2016keep}. And now citing a book reference \cite{bogo2016keep}. Please use the command  for the following MDPI journals, which use author-date citation: Administrative Sciences, Arts, Econometrics, Economies, Genealogy, Humanities, IJFS, JRFM, Languages, Laws, Religions, Risks, Social Sciences.

\section{Related Work}\label{sec:relatedwork}
The work presented in this section is closely related to our work and involves avoiding self-intersection with triangular mesh.

In~computer graphics, it is common to use proxy geometries to prevent self-intersection~\mbox{\cite{ericson2004real, thiery2013sphere}}. In~computer vision, recent works followed this approach to prevent self-intersection of 3D reconstruction results represented as triangular meshes. Sminchisescu et al. \cite{sminchisescu2001covariance} defined an implicit surface as a approximation of body shape to avoid self-intersection. Pons et al. \cite{pons2015metric} applied a set of spheres to approximate the interior volume of body mesh, and used the radius of each sphere to define a penalty term of self-intersection. These approaches are not accurate since the shape of human body can not be described exactly by spheres. To improve the accuracy of this approach, \mbox{Bogo et al.}~\cite{bogo2016keep} trained a regressor to generate capsules with minimum error to the body surface, then the authors further defined the penalty term as a 3D isotropic Gaussian derived from the capsule radius. It is worth mentioning that these approaches mentioned above do not strictly avoid self-intersection as approximations were applied to derive a differentiable penalty term. In \cite{wang2018pixel2mesh} the authors employed a Laplacian term to prevent the vertices from moving too freely, this penalty term avoids self-intersection to some degree. However this method still does not strictly avoid self-intersection since the Laplacian term acts just like a surface smoothness term preventing the 3D mesh from deforming too much.

Our work differs from previous works by identifying that a differentiable self-intersection penalty term is not necessary and the gradients can be calculated manually. We demonstrated appealing results in 3D pose estimation based on a  \DIFdelbegin \DIFdel{statical }\DIFdelend \DIFaddbegin \DIFadd{statistical }\DIFaddend body shape model.

\section{Self-intersection Penalty Term}\label{sec:method}
In this section, the details of our proposed self-intersection penalty term are discussed. We employed the SMPL human body shape model \cite{loper2015smpl} to evaluate our method. Essential \DIFdelbegin \DIFdel{notation is }\DIFdelend \DIFaddbegin \DIFadd{notations are }\DIFaddend provided here. SMPL model defines a function $\mathcal{M}(\bm{\beta},\bm{\theta};\bm{\Phi})$, where $\bm{\theta$} are the pose parameters, $\bm{\beta}$~are the shape parameters and $\bm{\Phi}$ are fixed parameters of the model. The output of this function is a triangular mesh $\bm{P}\in \mathbb{R}^{N\times3}$ with $N=6890$ vertices $P_i\in \mathbb{R}^3(i=1,\dots,N)$. The shape parameters $\bm{\beta}$ represents the coefficients of linear combination of a low number of principal body shapes learned from a dataset containing body scans \cite{robinette2002civilian}.
\subsection{Definition and Description}
Our method generally applies to meshes satisfying the conditions described below:
\begin{itemize}[leftmargin=*,labelsep=5.8mm]
\item	The mesh is a two-dimensional manifold.
\item   The mesh describes an orientable surface.
\item	The mesh is a closed surface.
\end{itemize}

Since the two-dimensional manifolds do not have to be connected, we can say a mesh with several disconnected parts also satisfies the conditions above. We demonstrated that our proposed method also works with a disconnected mesh in Section \ref{sec:experiment}.

To remove the self-intersection, the triangle mesh should be iteratively deformed by moving each vertex in a specific direction. The moving directions of vertices are obtained by computing the partial derivatives of the self-intersection penalty term which is defined as the volume of the self-intersection region in this paper. This penalty term is denoted as $E_{SPT}(V)$, where $V$ is \DIFaddbegin \DIFadd{the }\DIFaddend coordinates of all the vertices. An ideal self-intersection penalty term should satisfy the following conditions:
\begin{itemize}[leftmargin=*,labelsep=5.8mm]
\item	When there is no self-intersection, both the penalty term and the gradient of the penalty term should be \DIFdelbegin \DIFdel{0.
}\DIFdelend \DIFaddbegin \DIFadd{zero.
}\DIFaddend \item	When there is self-intersection, the value of penalty term indicates the degree of intersection.
\item   When there is self-intersection, the gradient of penalty term offers meaningful direction for~optimization.
\end{itemize}

Leaving the strategy of computing the value of penalty term aside, the method of computing gradient is discussed first.

\begin{figure}[H]
\centering
\DIFdelbeginFL %DIFDELCMD < \subfigure[]{
%DIFDELCMD < \begin{minipage}[t]{0.4\linewidth}
%DIFDELCMD < \centering
%DIFDELCMD < \includegraphics[width=5 cm]{figures/outintersection}
%DIFDELCMD < \end{minipage}
%DIFDELCMD < }
%DIFDELCMD < \subfigure[]{
%DIFDELCMD < \begin{minipage}[t]{0.4\linewidth}
%DIFDELCMD < \centering
%DIFDELCMD < \includegraphics[width=5 cm]{figures/inintersection}
%DIFDELCMD < \end{minipage}
%DIFDELCMD < }
%DIFDELCMD < %%%
\DIFdelendFL \DIFaddbeginFL \subfigure[]{
\begin{minipage}[t]{0.8\linewidth}
\centering
\includegraphics[width=14 cm,trim={2cm 8cm 1cm 8cm},clip]{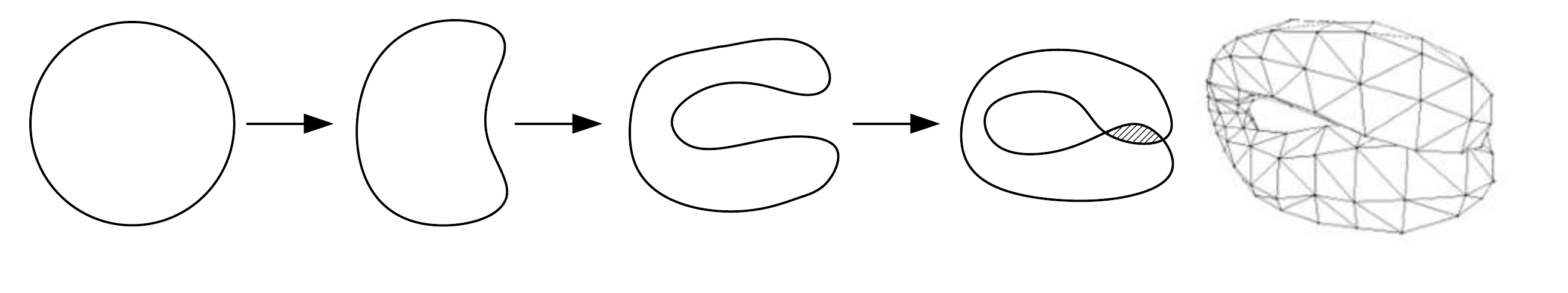}
\end{minipage}
}

\subfigure[]{
\begin{minipage}[t]{0.8\linewidth}
\centering
\includegraphics[width=14 cm,trim={2cm 8cm 1cm 8cm},clip]{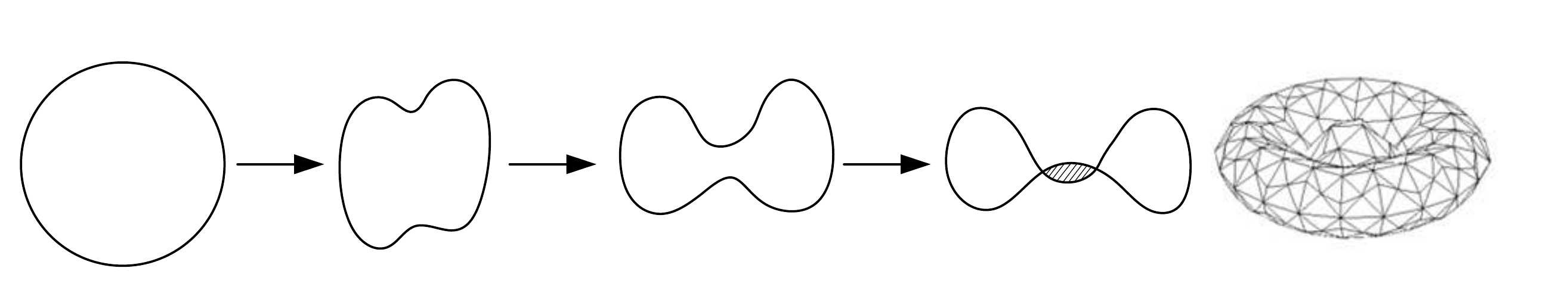}
\end{minipage}
}
\DIFaddendFL \caption{Two typical way of self-intersection: (a) Self-intersection due to interpenetration of the outer surface. (b) Self-intersection due to interpenetration of the inner surface.}
\label{fig:intersection}
\end{figure}

The first step of computing the partial derivatives is separating the vertices into two sets: (1)~vertices in the self-intersection region and (2) %Bold font removed from all, please confirm meaning is retained.
vertices not in the self-intersection region. We implemented this separation by emitting a beam of detection rays, the density of rays \DIFdelbegin \DIFdel{depends on }\DIFdelend \DIFaddbegin \DIFadd{is manually set according to }\DIFaddend the number of triangles in the mesh. \DIFaddbegin \DIFadd{An appropriate setting of density of rays guarantees the accuracy of classification and low memory consumption. }\DIFaddend There are two typical ways of intersection shown in Figure \ref{fig:intersection}. According to the type of self-intersection, the set of vertices in self-intersection could be further separated into two sets. Overall, the vertices are divided into three sets: (a) vertices in the self-intersection region due to interpenetration of the outer surface, denoted as $V_{out}$;  (b) vertices in the self-intersection region due to interpenetration of the inner surface, denoted as $V_{in}$  and (c) vertices not in the self-intersection region, denoted as $V_0$;. Based on the classification result, the partial derivative of the penalty term with respect to coordinate of a vertex can be obtained according to the normal vector and which set the vertex is belonging to.

It is worth pointing out that self-intersection described in Figure \ref{fig:intersection} (b) rarely occurs for \DIFdelbegin \DIFdel{statical }\DIFdelend \DIFaddbegin \DIFadd{statistical }\DIFaddend body shape model. To maintain the generality of our method, both of these two type of self-intersection are considered in the following discussion.

\subsection{Detection and Classification of Vertices}

\begin{figure}[H]
\centering
\includegraphics[width=.8\textwidth]{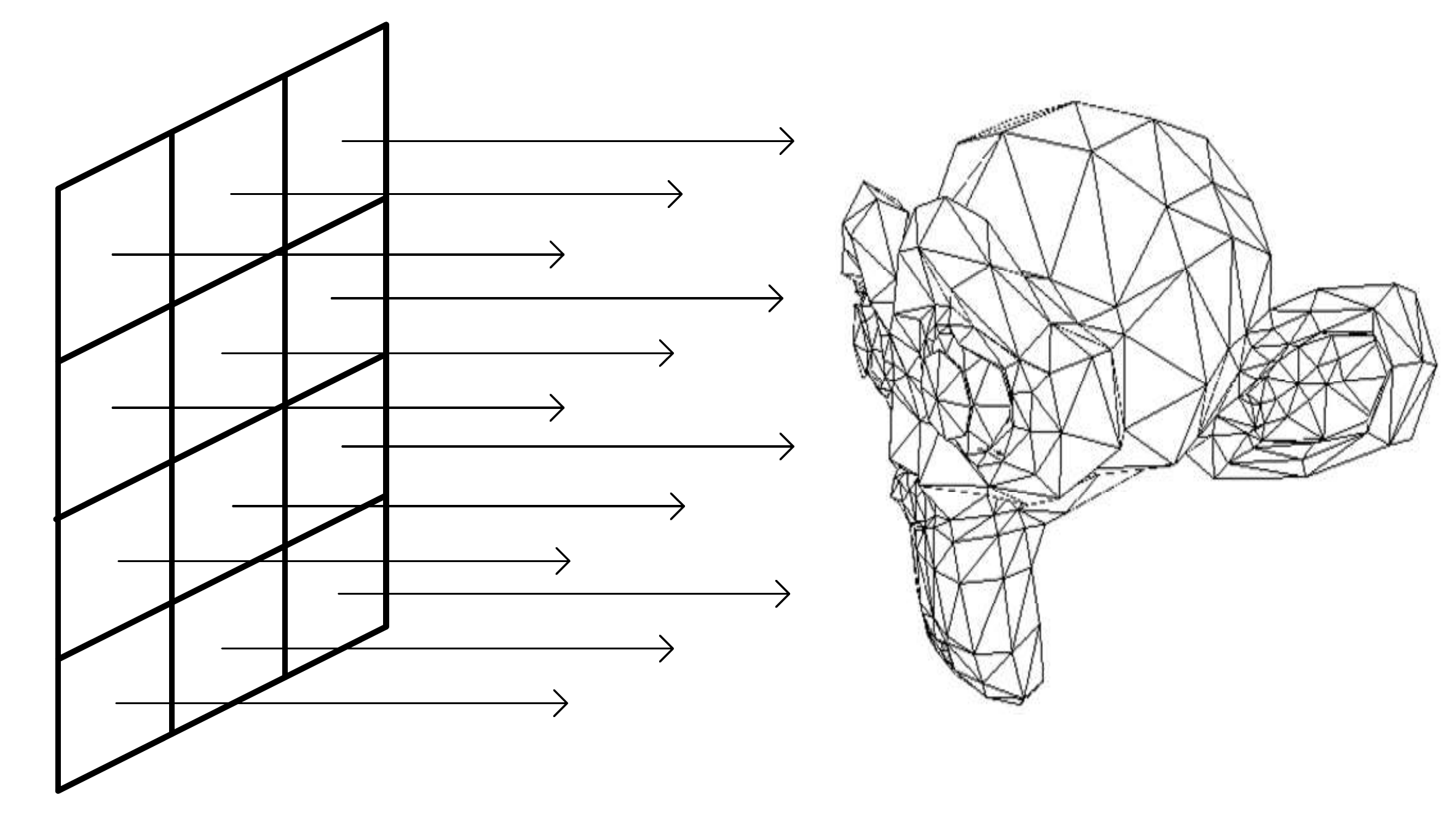}
\caption{Schematic representation of our approaches to detect self-intersection (the image of triangular mesh is rendered by Blender).}
\label{fig:detect}
\end{figure}

\begin{figure}[H]
\centering
\includegraphics[width=.8\textwidth]{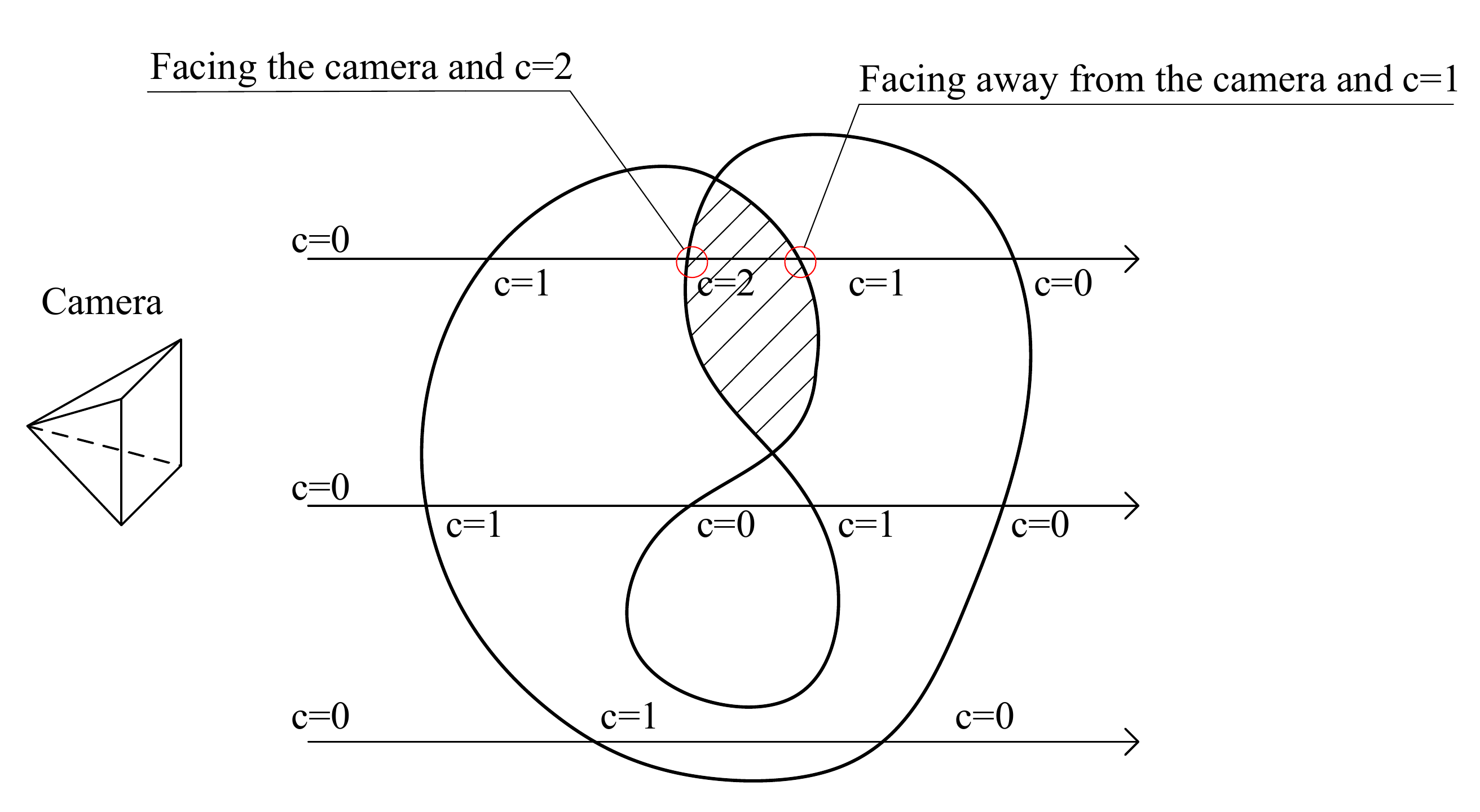}
\caption{\DIFdelbeginFL \DIFdelFL{A }\DIFdelendFL \DIFaddbeginFL \DIFaddFL{An }\DIFaddendFL intuitive representation of our algorithm to detect self-intersection. The detection rays are emitted from the camera, the dashed area represents the self-intersection region. The vertices in red circle were detected to be in self-intersection according to the counter and orientation.}
\label{fig:algorithm}
\end{figure}

To compute the partial derivatives of our proposed self-intersection penalty term, it is necessary to divide the vertices into three sets: $V_0$, $V_{in}$ and $V_{out}$. A camera screen with pixels arranged in a square with $H$ rows and $W$ columns is set in front of the 3D mesh such that the orthogonal projection of the 3D mesh falls totally inside the screen. \DIFdelbegin \DIFdel{Detection rays emit }\DIFdelend \DIFaddbegin \DIFadd{It is worth noting that the camera mentioned here is used only to emit detection rays, not for rendering and visualization. Detection rays are emitted }\DIFaddend from the center of each pixel to detect self-intersection, as is shown in Figure \ref{fig:detect}. The detail of the detection and classification is presented in Algorithm \ref{alg:detect}. To make our algorithm more intuitive, a schematic representation is given in Figure \ref{fig:algorithm}.

\begin{algorithm}[H]
\begin{spacing}{1.0}
  \caption{Illustration of self-intersection detection and classification of vertices}
  \LinesNumbered
  \label{alg:detect}
   \KwIn{the height $H$ and width $W$ of the screen, $V$ containing coordinates of all vertices, $F$ containing vertex indexes of all triangles.}
   \KwOut{$V_0$ containing vertices not in self-intersection, $V_{out}$ containing vertices in self-intersection of the outer surface, $V_{in}$ containing vertices in self-intersection of the inner surface.}
    \textbf{initialize} $V_0$, $V_{out}$ and $V_{in}$ as empty sets\\
    \textbf{initialize} the linked list of each pixel as empty list\\
    \For{each triangle in $F$}
    {
       project the triangle onto the screen by orthogonal projection\\
       \For{each pixel in the screen}
       {
          \If{this pixel located inside the triangle}
          {
             insert this triangle into the linked list of this pixel in the order of smallest to largest in terms of depth
          }
       }
    }
    \For{each pixel in the screen}
    {
        \If{the linked list of this pixel is empty}
        {
            \textbf{continue}
        }
        \textbf{initialize} $counter\gets0$ \\
        \For{each triangle in the linked list}
        {
            \If{this triangle is facing the camera}
            {
                $count\gets counter+1$
            }
            \Else
            {
                $count\gets counter-1$
            }
            \If{this triangle is facing the camera}
                {
                    \If{$counter=1$}
                    {
                        append the three vertices of this triangle to $V_0$
                    }
                    \ElseIf{$counter=2$}
                    {
                        append the three vertices of this triangle to $V_{out}$
                    }
                    \ElseIf{$counter=0$}
                    {
                        append the three vertices of this triangle to $V_{in}$
                    }
                }
                \Else
                {
                    \If{$counter=0$}
                    {
                        append the three vertices of this triangle to $V_0$
                    }
                    \ElseIf{$counter=1$}
                    {
                        append the three vertices of this triangle to $V_{out}$
                    }
                    \ElseIf{$counter=-1$}
                    {
                        append the three vertices of this triangle to $V_{in}$
                    }
                }
        }
    }
    \Return $V_0$,$V_{out}$ and $V_{in}$
  \end{spacing}
\end{algorithm}

\subsection{Gradients Calculation}
The self-intersection penalty term $E_{SPT}(V)$ is defined as:
\begin{equation}
E_{SPT}(V)=V_{intersect}
\end{equation}
where $V$ denotes the set of coordinates of all vertices, $V_{intersection}$ is the volume of the self-intersection region. For the sake of further discussion, a vertex $p_t (t=1,\dots,N)$ is randomly chosen from the triangular mesh with $N$ vertices. Coordinates of all vertices are frozen, except for $p_t$. The coordinate of $p_t$ is denoted as $(x_t, y_t, z_t)$.

Under the preceding assumptions, the penalty term \DIFdelbegin \DIFdel{$E_p(V)$ }\DIFdelend \DIFaddbegin \DIFadd{$E_{SPT}(V)$ }\DIFaddend can be regarded as a function of $x_t$,$y_t$ and \mbox{$z_t$}.
\begin{equation}
E_{SPT}(V)=f(x_t,y_t,z_t)
\end{equation}

To compute $\frac{\partial E_{SPT}(V)}{\partial x_t}$, $\frac{\partial E_{SPT}(V)}{\partial y_t}$ and $\frac{\partial E_{SPT}(V)}{\partial z_t}$, $p_t$ is displaced from $(x_t, y_t, z_t)$ to $(x_t+\Delta x,y_t+\Delta y,$ $z_t+\Delta z)$. Then the change in $E_{STP}(V)$ due to the displacement could be represented as:
\begin{equation}
\Delta E_{SPT}(V)=f(x_t+\Delta x, y_t+\Delta y, z_t+\Delta z)-f(x_t, y_t, z_t)
\end{equation}

Further computation is hard to continue without imposing constraints on $p_t$. First, for the simplest case, $p_t$ is assumed to belong to $V_0$, that is to say $p_t \in V_0$. Since tiny displacement of $p_t$ brings no effect to the volume of self-intersection, it is obvious that:
\begin{equation}
\Delta E_{SPT}(V)=\Delta f=0, p_t \in V_0
\end{equation}

The gradient of $E_{SPT}(V)$ could be represented as:
\begin{equation}
\left(
  \begin{array}{c}
    \frac{\partial E_{SPT}(V)}{\partial x_t} \\[0.5ex]
    \frac{\partial E_{SPT}(V)}{\partial y_t} \\[0.5ex]
    \frac{\partial E_{SPT}(V)}{\partial z_t} \\
  \end{array}
\right)
=
\left(
  \begin{array}{c}
    \lim_{\Delta x \to 0}\frac{\Delta f}{\Delta x} \\[0.5ex]
    \lim_{\Delta y \to 0}\frac{\Delta f}{\Delta y} \\[0.5ex]
    \lim_{\Delta z \to 0}\frac{\Delta f}{\Delta z} \\
  \end{array}
\right)
=
\left(
  \begin{array}{c}
    0 \\
    0 \\
    0 \\
  \end{array}
\right),
p_t \in V_0
\end{equation}

\begin{figure}[H]
\centering
\includegraphics[width=8 cm]{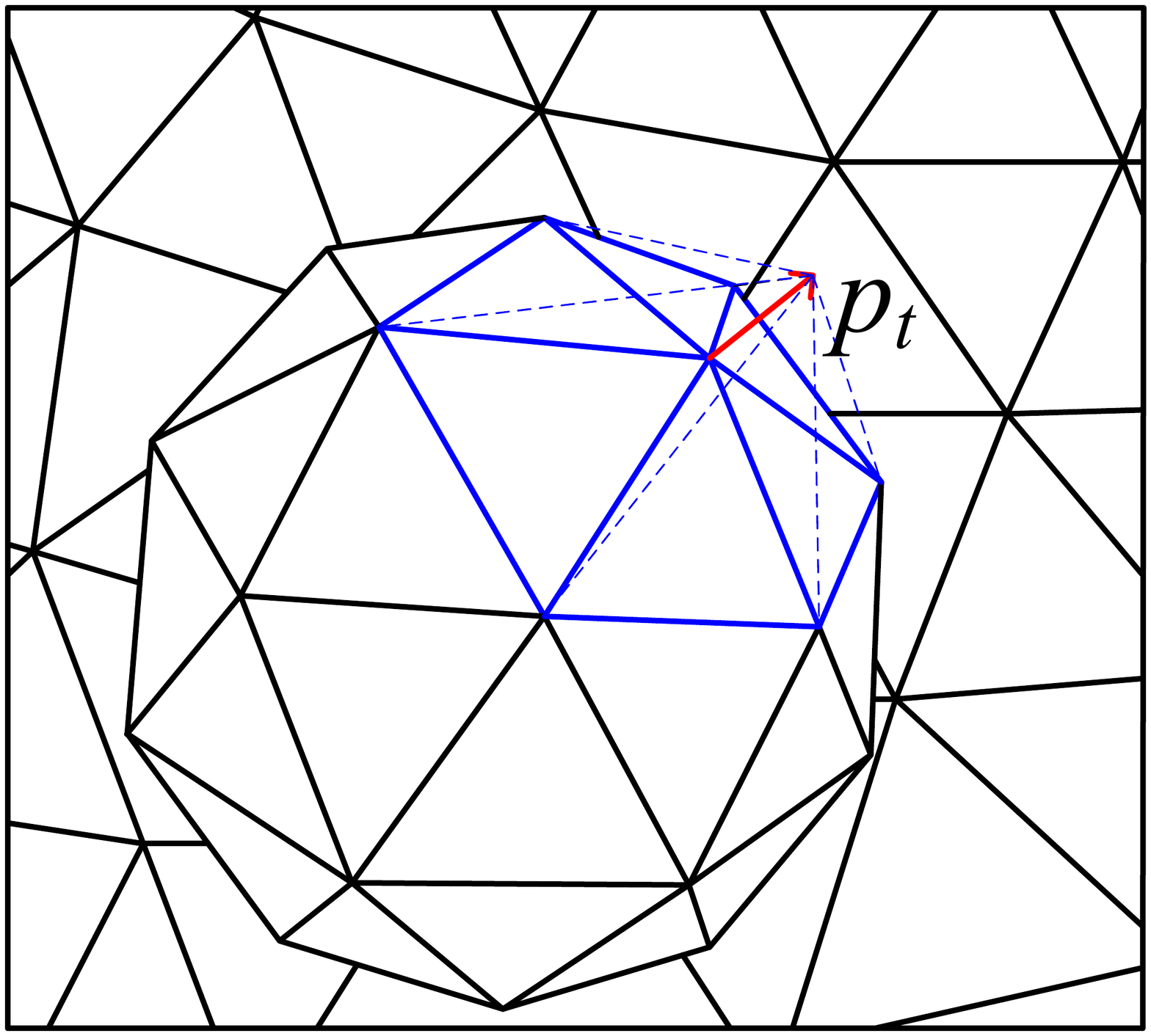}
\caption{A tiny displacement on the vertex $p_t$. The change in the volume of self-intersection is equal to the sum of volume of several neighboring tetrahedrons.}
\label{fig:gradient}
\end{figure}

For the second case $p_j \in V_{out}$, more assumptions need to be made for a detailed discussion. We~assume that there are $n$ neighboring triangles sharing $p_t$ as a common vertex. One of the neighboring triangles is denoted as $T_l(l=1,\dots,n)$, the area of $T_l$ is represented as $S_l$ and the unit normal vector of $T_l$ is denoted as $\bm{n}_l$. An intuitive representation of this situation is shown in Figure \ref{fig:gradient}.

The change in the volume of self-intersection due to tiny displacement of vertex $p_t$ can be represented~as:
\begin{equation}
\Delta f=\Delta V_{intersection}=\frac{1}{3}\sum_{l=1}^nS_l\bm{n}_l(\Delta x\bm{i}+\Delta y\bm{j}+\Delta z\bm{k}),p_t \in V_{out}
\end{equation}
where $\bm{i}$,$\bm{j}$ and $\bm{k}$ are unit vectors in the same directions as the positive directions of $x$, $y$ and $z$ axes.

The gradient can be obtained as:
\begin{equation}
\left(
  \begin{array}{c}
    \frac{\partial E_{SPT}(V)}{\partial x_t} \\[0.5ex]
    \frac{\partial E_{SPT}(V)}{\partial y_t} \\[0.5ex]
    \frac{\partial E_{SPT}(V)}{\partial z_t} \\
  \end{array}
\right)
=
\left(
  \begin{array}{c}
    \lim_{\Delta x \to 0}\frac{\Delta f}{\Delta x} \\[0.5ex]
    \lim_{\Delta y\to 0}\frac{\Delta f}{\Delta y} \\[0.5ex]
    \lim_{\Delta z\to 0}\frac{\Delta f}{\Delta z} \\
  \end{array}
\right)=
\left(
  \begin{array}{c}
    \frac{1}{3}\sum_{l=1}^nS_l\bm{n}_l\bm{i} \\[0.5ex]
    \frac{1}{3}\sum_{l=1}^nS_l\bm{n}_l\bm{j} \\[0.5ex]
    \frac{1}{3}\sum_{l=1}^nS_l\bm{n}_l\bm{k} \\
  \end{array}
\right),p_t \in V_{out}
\end{equation}

The equation above can be simplified as:
\begin{equation}
\left(
  \begin{array}{c}
    \frac{\partial E_{SPT}(V)}{\partial x_t} \\[0.5ex]
    \frac{\partial E_{SPT}(V)}{\partial y_t} \\[0.5ex]
    \frac{\partial E_{SPT}(V)}{\partial z_t} \\
  \end{array}
\right)=
\frac{1}{3}\sum_{l=1}^nS_l\bm{n}_l,p_t \in V_{out}
\end{equation}

For the last case $p_t\in V_in$, the derivation process of partial derivative is similar to the process described in the second case, and the results are same in magnitude but opposite in sign. The gradient in this case can be obtained as:
\begin{equation}
\nabla E_{SPT}(V)=-\frac{1}{3}\sum_{l=1}^nS_l\bm{n}_l,p_t \in V_{in}
\end{equation}

In summary, the gradient of self-intersection penalty term with respect to the coordinate of vertex $p_t$ can be represented as: %Is the bold of number 0 in Equation (10) necessary? If not, please remove.
\begin{equation}
\nabla E_{SPT}(V)=
\begin{cases}
\vec{0}&p_t\in V_0\\
\frac{1}{3}\sum_{l=1}^nS_l\bm{n}_l&p_t \in V_{out}\\
-\frac{1}{3}\sum_{l=1}^nS_l\bm{n}_l&p_t \in V_{in}
\end{cases}
\end{equation}

Employing the equation above for a gradient-based optimization algorithm to remove self-intersection works well in most cases. But when there are great differences between the areas of triangles, the process of optimization tends to be unstable. To solve this problem, a modified version of the gradient formula is presented:%Is the bold of number 0 in Equation (10) necessary? If not, please remove.
 \begin{equation}
\nabla E_{SPT}(V)=
\begin{cases}
\vec{0}&p_t\in V_0\\[0.5ex]
\frac{\sum_{l=1}^nS_l\bm{n}_l}{\left\|\sum_{l=1}^nS_l\bm{n}_l\right\|_2}&p_t \in V_{out}\\[1ex]
-\frac{\sum_{l=1}^nS_l\bm{n}_l}{\left\|\sum_{l=1}^nS_l\bm{n}_l\right\|_2}&p_t \in V_{in}
\end{cases}
\label{eq:normalize}
\end{equation}

For the value of $E_{SPT}(V)$, it is difficult and unnecessary to calculate the exact volume of the intersection region by coordinates of vertices. Instead, $E_{SPT}(V)$ is assigned with the ratio of the number of vertices in self-intersection to the number of vertices not in self-intersection. This ratio is easy to calculate and significant for indicating the degree of self-intersection. The expression of $E_{SPT}(V)$ is presented as:
\begin{equation}
\begin{aligned}
E_{SPT}(V)&=\frac{\vert V_{out}\vert+\vert V_{in}\vert}{\vert V_{out}\vert+\vert V_{in}\vert+\vert V_0\vert}\\&= \frac{\vert V_{out}\vert+\vert V_{in}\vert}{N}
\end{aligned}
\end{equation}
where $\vert\cdot\vert$ denotes the number of elements in a set.

So far, the forward and backward processes of our proposed self-intersection penalty term have been defined.

\section{Experimental Results and Discussion}\label{sec:experiment}

In this section, experiments were conducted to evaluated the effectiveness of our proposed self-intersection penalty term (SPT). We employed a statistical body shape model SMPL \cite{loper2015smpl} to show the effectiveness of our method.

\newpage

\subsection{Self-Intersection Removal on a Single SMPL Model}
To evaluate the validity of our method, we set the pose parameters $\bm{\theta}$ to generate triangular meshes with self-intersection deliberately. Then we performed gradient descent algorithm to optimize the pose parameters $\bm{\theta}$ to remove the self-intersection, the shape parameters $\bm{\beta}$ were \DIFaddbegin \DIFadd{the ground truth values from dataset and were }\DIFaddend fixed during iterations. The learning rate of gradient descent was set to $1.0\times10^{-4}$. The number of detection rays was set to $512\times512$.

\begin{figure}[H]
\centering
\includegraphics[width=14 cm,trim={0 2cm 0 1cm},clip]{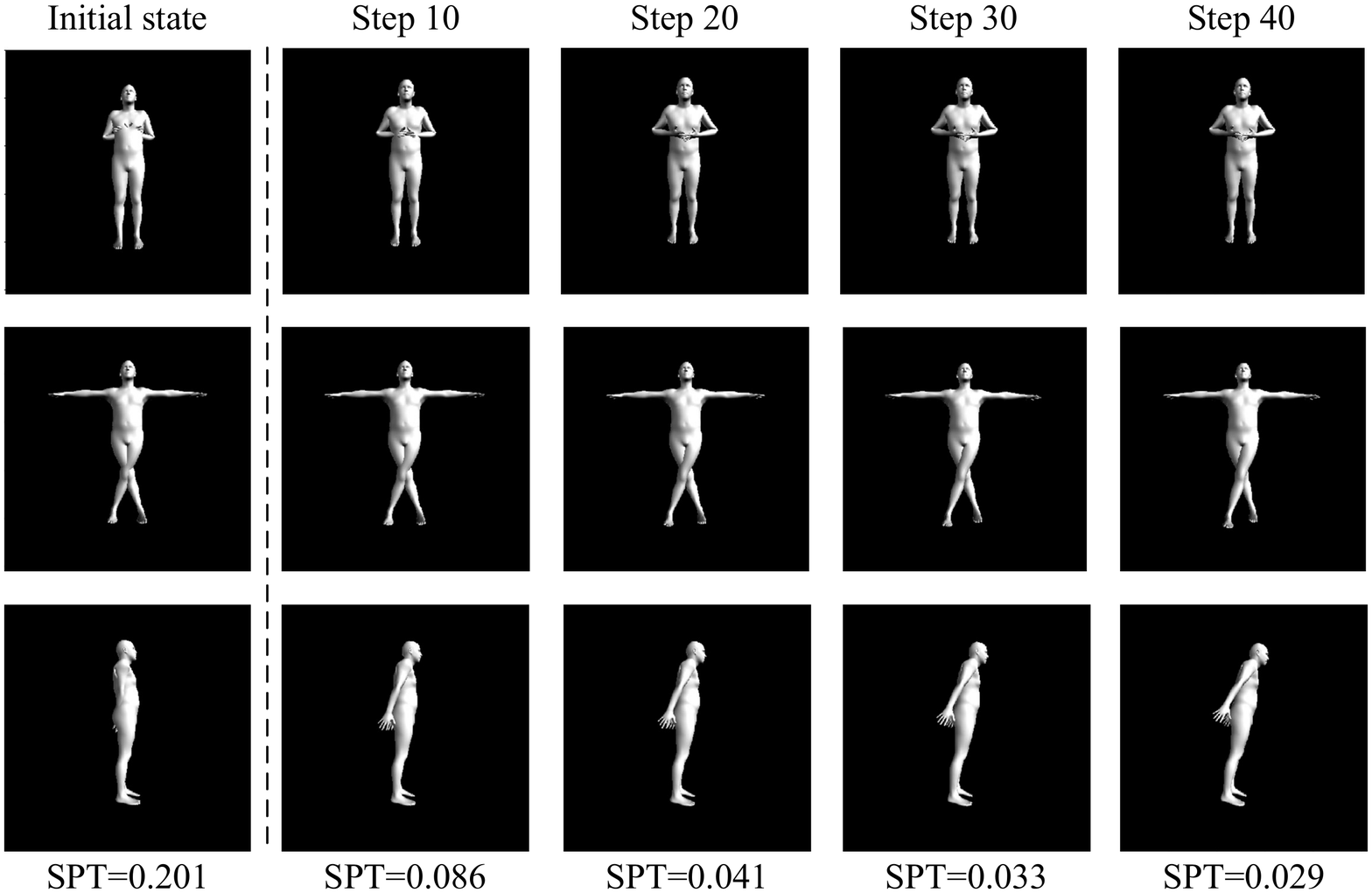}
\caption{Images rendered from the iterative process. First column: Images rendered from initial meshes with self-intersection. Second through fourth columns: Images rendered from optimized meshes every 10 iterations. The self-intersection penalty term (SPT) values in the bottom of each column denote the average SPT of three SMPL %Please define if appropriate.
models. }
\label{fig:iters}
\end{figure}

The process of iteration is visualized in Figure \ref{fig:iters}. \DIFdelbegin \DIFdel{Images in first column are the initial states of iteration. Images in second through fourth columns are the intermediate results after every 10 iterations. }\DIFdelend We can see that the gradient calculated via proposed method works effectively in a gradient descent algorithm to minimize the \DIFdelbegin \DIFdel{SPT}\DIFdelend \DIFaddbegin \DIFadd{value of SPT, that is to say, minimize the number of vertices in self-intersection region}\DIFaddend .

\begin{figure}[H]
\centering
\includegraphics[width=15 cm,trim={0 2cm 0 3cm},clip]{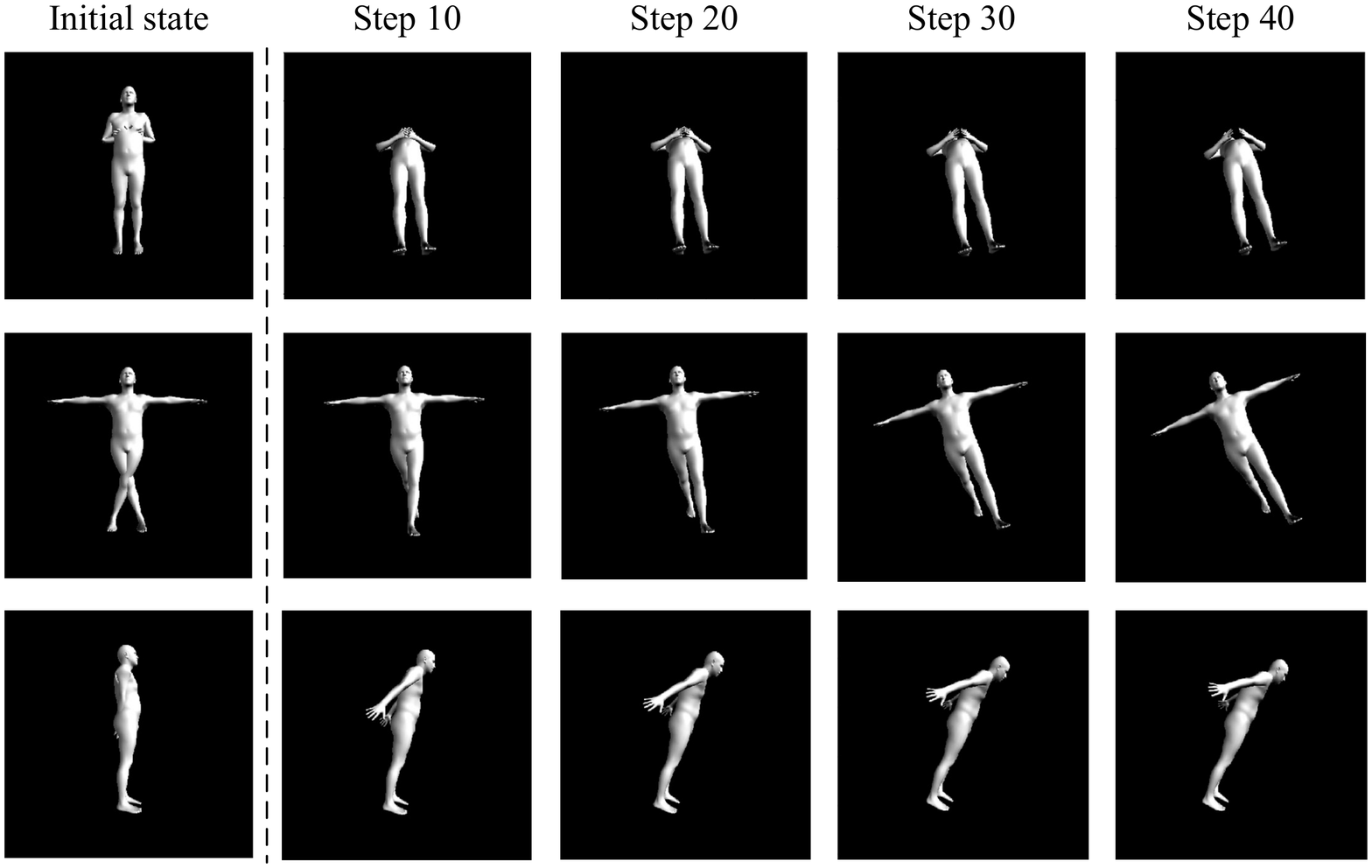}
\caption{Images rendered from the iterative process without gradient normalization. First column: images rendered from initial meshes with self-intersection. Second through fourth columns: images rendered from optimized meshes every 10 steps.}
\label{fig:cmp}
\end{figure}

In order to show the necessity of \DIFaddbegin \DIFadd{gradient }\DIFaddend normalization which is presented in Equation \eqref{eq:normalize}, an experiment with same conditions as the experiment described above, but without gradient normalization, was conducted. The visualized result is shown in Figure \ref{fig:cmp}. Since there are great differences between the areas of triangles in the body meshes, gradients of vertices differ greatly in magnitudes. This often leads to unstable iterations and unpredictable results \DIFaddbegin \DIFadd{and we demonstrated that this problem can be solved by gradient normalization}\DIFaddend .

As can be seen from Figures \ref{fig:iters} and \ref{fig:cmp}, gradient normalization improves the stability of optimization. Therefore it can be concluded that gradient normalization is significant for 3D pose estimation since unstable iterations often lead to failure predictions.

\subsection{Intersection and Self-Intersection Removal on Multiple SMPL Models}

\begin{figure}[H]
\centering
\includegraphics[width=16 cm,trim={0 2cm 0 2cm},clip]{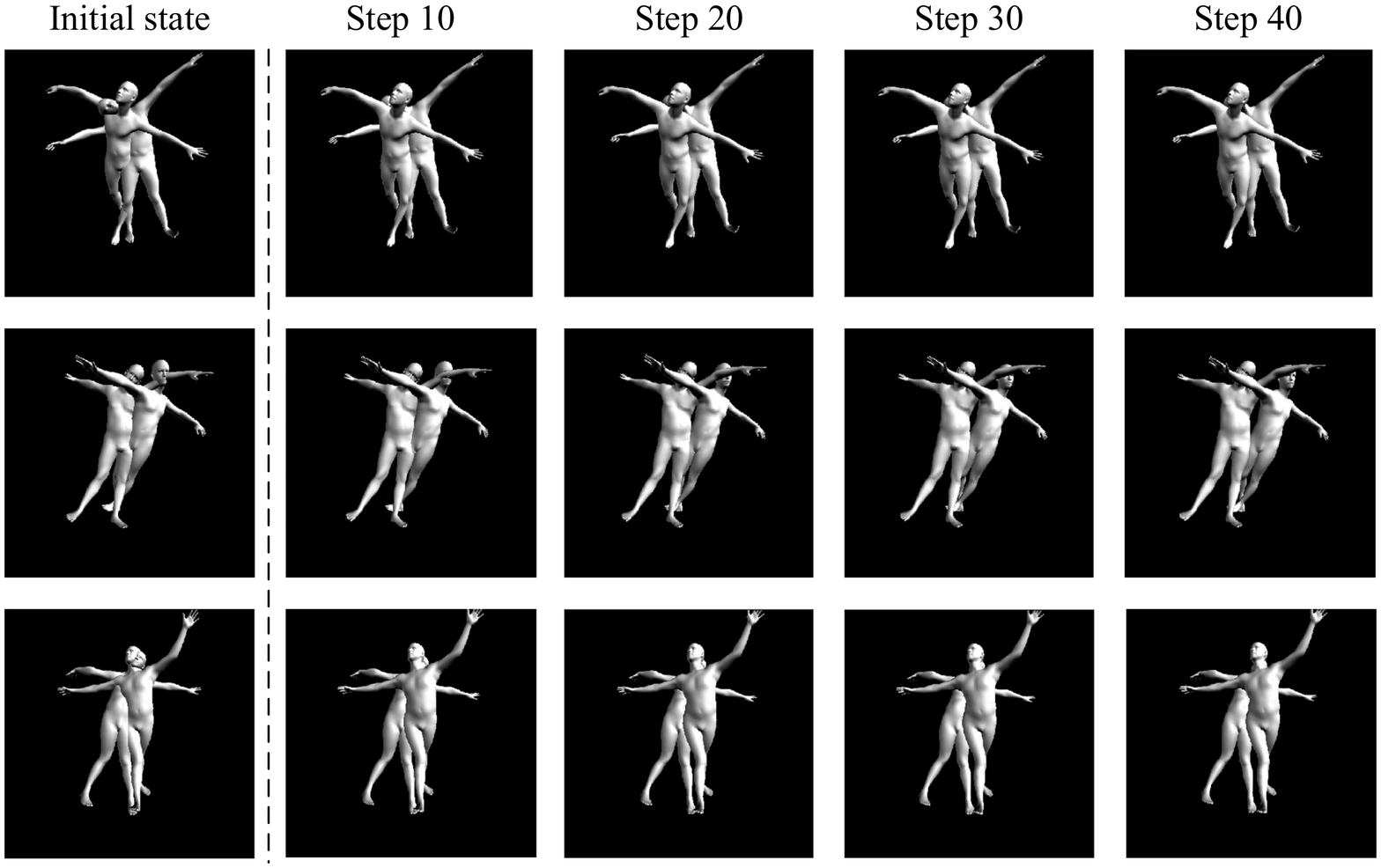}
\caption{Images rendered from the iterative process with two SMPL models. First column: Images rendered from initial meshes with self-intersection. Second through fourth columns: Images rendered from optimized meshes every 10 steps.}
\label{fig:mul}
\end{figure}

To demonstrate that our method applies to general closed surfaces, an experiment on mesh with two disconnected surfaces was carried out. In the experiment, two SMPL mesh models were generated and were regarded as one mesh, gradient descent was employed to remove the intersection between the two SMPL models and the self-intersection of themselves. We visualized the result in Figure \ref{fig:mul}.

Figure \ref{fig:mul} shows that our proposed self-intersection penalty term can remove both the intersection between different body meshes and the self-intersection of each body mesh. This property of our method is of great significance for multi-person 3D pose estimation.

\subsection{3D Pose Estimation from 2D Joints}
We tested our proposed self-intersection penalty term (SPT) on \DIFdelbegin \DIFdel{\hl{UP3D}
 }\DIFdelend \DIFaddbegin \DIFadd{UP-3D %Please define if appropriate.
  }\DIFaddend \cite{lassner2017unite} whose sample images were labeled with ground truth 2D joints. To estimate 3D pose from 2D joints, we defined an objective function as:
\begin{equation}
E(\bm{\theta})=E_J(\bm{\theta},\bm{\beta},K,J_{gt})+E_{SPT}(\mathcal{M}(\bm{\beta},\bm{\theta};\bm{\Phi}))
\label{eq:objective}
\end{equation}
where $K$ are camera parameters and $J_{gt}$ is the ground truth of 2D joints. $E_J$ represents the error between projected joints of the SMPL model and the ground truth 2D joints.

The shape parameters $\bm{\beta}$ are fixed during optimization.
Since the objective function defined in Equation \eqref{eq:objective} is differentiable, gradient descent can be directly applied to optimize the pose parameters $\bm{\theta}$ by minimizing the objective function. To demonstrate that SPT can improve the accuracy of 3D pose estimation by excluding unreasonable predictions, we also performed the optimization of objective function without SPT which can be represented as:
\begin{equation}
E'(\bm{\theta})=E_J(\bm{\theta},\bm{\beta},K,J_{gt})
\label{eq:objective1}
\end{equation}

\begin{figure}[H]
\centering
\includegraphics[width=.96\textwidth]{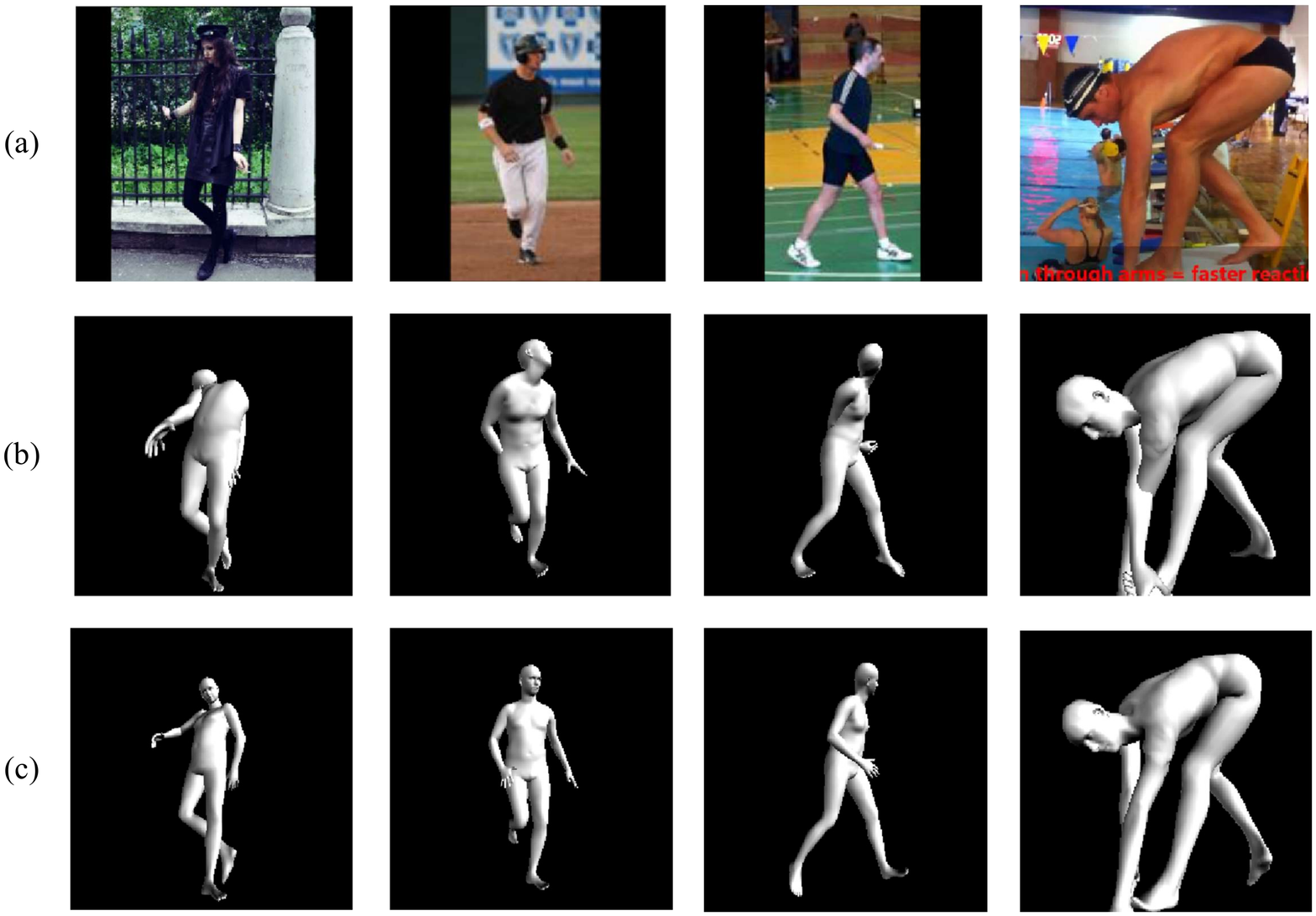}
\caption{Visualized results of 3D pose estimation. Images in row (\textbf{a}) are the input images, images in row (\textbf{b}) are fitting results without SPT, images in row (\textbf{c}) are fitting results with SPT.}
\label{fig:pose}
\end{figure}

Figure \ref{fig:pose} visually compares the results of two different objective functions on a few images from \DIFdelbegin \DIFdel{UP3D }\DIFdelend \DIFaddbegin \DIFadd{UP-3D }\DIFaddend dataset. It is obvious that minimizing the error between projected joints and the ground truth 2D joints directly without self-intersection penalty term tends to result in body meshes with self-intersection. The fitting results with SPT are more natural and more reasonable compared with the results without SPT. This experiment demonstrated that it is effective to add our proposed SPT into the objective function to avoid self-intersection of body meshes in optimization-based 3D pose estimation.

\DIFaddbegin \subsection{Comparison with State-of-the-Art}
\DIFadd{To compare our proposed method with other state-of-the-art methods, we conducted an experiment on UP-3D dataset. Since our iterative optimization-based method often fails because of local minima, we used a reduced test set of 139 images selected by Tan et al. }\cite{tan2017indirect} \DIFadd{to limit the range for the global rotation of body shape model.
}

\begin{figure}[H]
\centering
\includegraphics[width=16 cm,trim={0 7.5cm 0 7cm},clip]{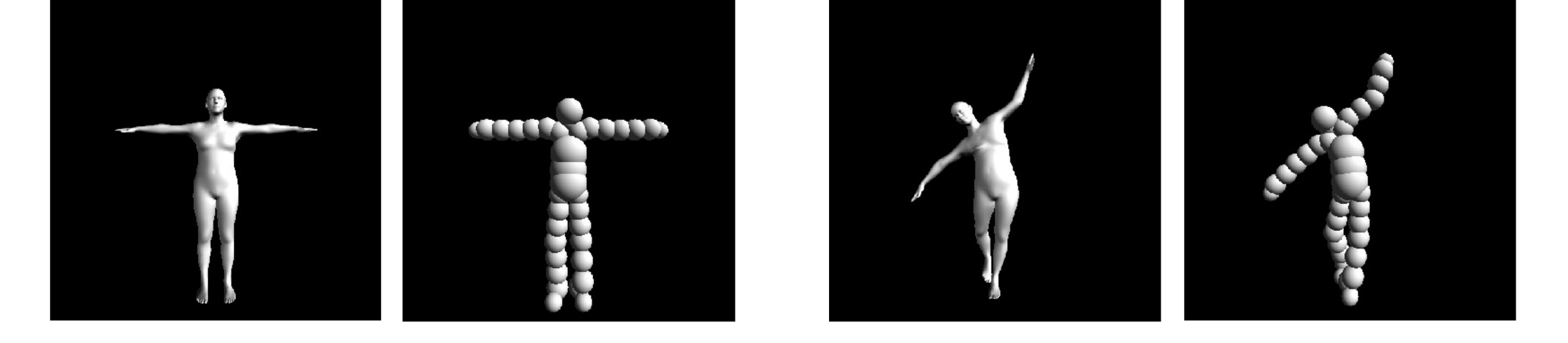}
\caption{\DIFaddFL{Spheres designed to approximate the human body are kept in the same pose with the body shape model.}}
\label{fig:sphere_body}
\end{figure}

\DIFadd{We implemented two state-of-the-art methods for qualitative and quantitative comparisons. In~the baseline method, the objective function is defined as the reprojection error of 2D joints only.
These methods are described in more detail below:
}\begin{itemize}[leftmargin=*,labelsep=5.8mm]
\item{\textbf{Reprojection error of 2D joints (RE) only}}\\	\DIFadd{This method only minimizes the error between ground truth 2D joints and projected 2D joints to estimate the 3D pose.
}\item{\textbf{RE + Laplacian regularization (LR) }}\\\DIFadd{Laplacian regularization is proposed by Wang et al. }\cite{wang2018pixel2mesh} \DIFadd{to prevent the vertices from moving too freely and potentially avoids mesh self-intersection in triangular mesh based 3D reconstruction. We employed this method in 3D pose estimation and the objective function is defined as the sum of reprojection error of 2D joints and the Laplacian regularization term.
}\item{\textbf{RE + Sphere approximation (SA) }}\\	\DIFadd{Pons-Moll et al. }\cite{pons2015metric} \DIFadd{built a set of spheres as a coarse approximation to the body shape model and derived a differentiable penalty term via calculating the intersection between spheres. To~implement this method, We designed a set of spheres to approximate the surface of human body, as is shown in Figure \ref{fig:sphere_body} . The objective function is defined as the sum of reprojection error of 2D joints and the intersection between spheres.
}\item{\textbf{RE + SPT}}\\\DIFadd{This is our proposed method whose objective function is defined as the sum of the reprojection error of 2D joints and the self-intersection penalty term proposed in this paper.
}\end{itemize}

\subsubsection{Qualitative Comparison}
\begin{figure}[H]
\centering
\includegraphics[width=16 cm,trim={0 1cm 0 1cm},clip]{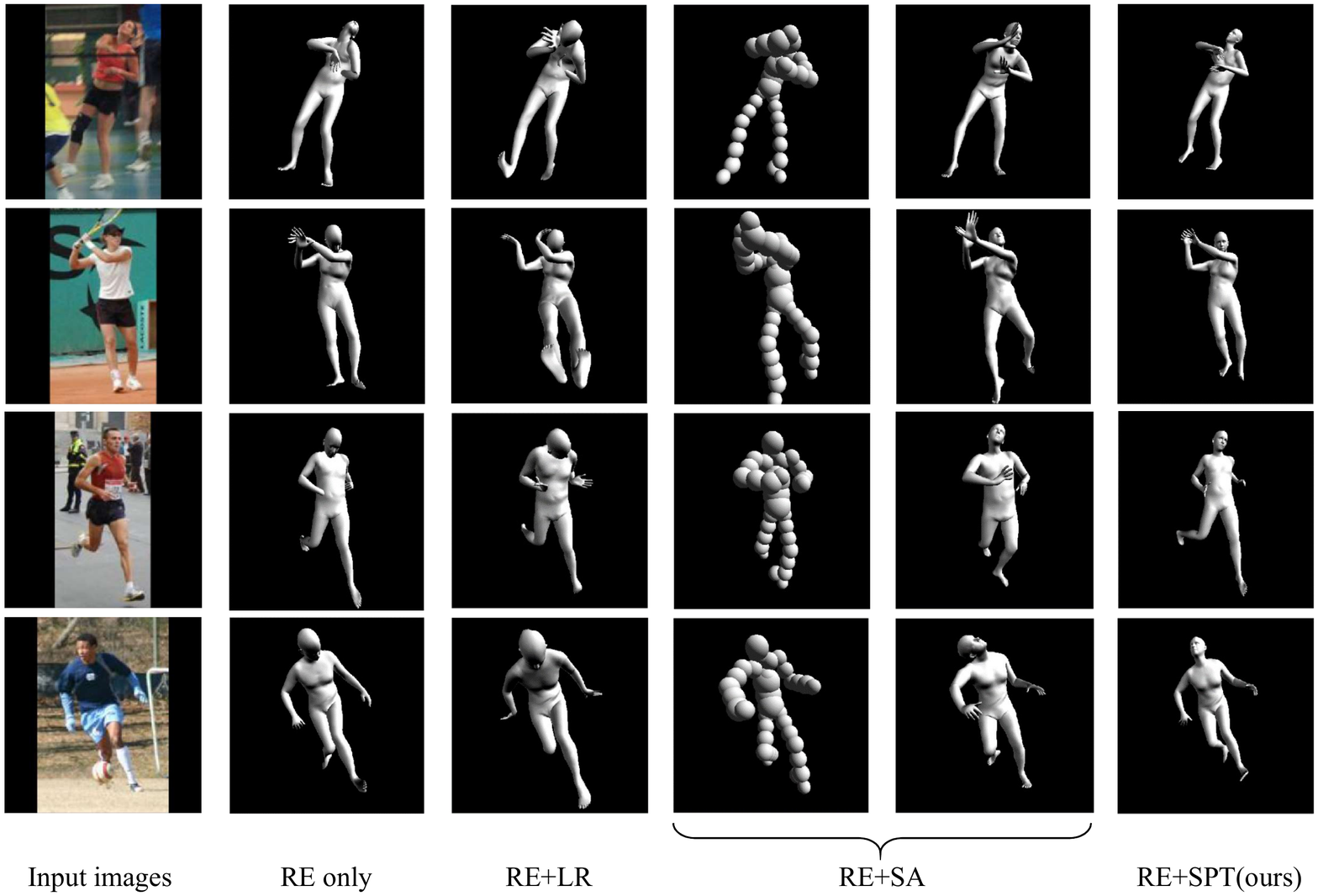}
\caption{\DIFaddFL{3D pose estimation results by four different methods. First column: Input images. Second column: Results obtained by minimizing reprojection (RE) error of 2D joints only. Third column: Results obtained by minimizing the sum of reprojection error and Laplacian regularization (LR). Fourth column through fifth column: Results obtained by minimizing the sun of reprojection error and the intersection between spheres. Sphere Approximation (SA). Sixth column: Results obtained by our proposed approach.}}
\label{fig:qualitative}
\end{figure}

\DIFadd{We implemented two state-of-the-art methods and one baseline method for qualitative comparison. Figure \ref{fig:qualitative} presents a part of results from the reduced test set of UP-3D by our proposed method and other methods. These results demonstrate that our proposed method can remove the self-intersection of the statistical body shape model effectively and produces more reasonable results.
}

\DIFadd{As is can seen from Figure \ref{fig:qualitative} , the results obtained by the baseline method without any self-intersection penalty tends to intersects with itself. The Laplacian regularization can not strictly avoid self-intersection and often leads to unnatural results. We can see that it is not suitable to employ Laplacian regularization in statistical body shape model because of the fact that this laplacian term brings negative effect to 3D pose estimation. The method of sphere approximation is very competitive in removing the self-intersection of body mesh, however method requires designing a appropriate set of spheres and we found that it is an excessive trivial procedure to set the radius and the coordinate of each sphere appropriately. In addition, since the body mesh can not be approximated accurately by spheres, this method may lead to fail results by simply removing the intersection between spheres.
}

\DIFadd{Compared with other approach, our proposed method can obtain more visually appealing results from two points. One is that when there is no self-intersection, our proposed SPT will have no effect on the body mesh, this means no side effects on 3D pose estimation. The other is that our proposed approach can avoid the self-intersection of body mesh strictly without taking any approximation.
}

\subsubsection{Quantitative Comparison}
\DIFadd{To the best of our knowledge, there is no commonly used evaluation metric for methods preventing self-intersection of triangular mesh.
 In order to compare our method with other state-of-the-art approaches quantitatively, we adopt the per vertex errors as 3D pose estimation metric and we used the percentage of vertices in region of self-intersection computed by our proposed algorithm to evaluate the performance of our method and other state-of-the-art approaches. It should be noted that it is inappropriate to evaluate these methods only by per vertex error or percentage of vertices in self-intersection because an ideal approach of avoiding self-intersection should achieve both minimum per vertex error and minimum percentage of vertices in self-intersection region. Therefore these two evaluation metrics, per vertex error and percentage of vertices in self-intersection, were adopted to conduct a quantitative comparison.
}

\DIFadd{The quantitative results of the baseline method, the other state-of-the-art approaches and our proposed method is shown in Table \ref{tab:quantitative}. The Laplacian regularization method achieved a lower percentage of vertices in self-intersection but a higher per vertex error compared to the baseline method, which demonstrates that the Laplacian regularization does work in avoiding self-intersection but the side effect leads to loss of precision in 3D pose estimation. The approach of approximating body shape by a set of spheres outperforms the baseline method both in per vertex error and percentage of vertices in self-intersection. It is undeniable that this method may perform better with more carefully designed spheres, but it will be extremely tedious to implement this method. Our proposed approach outperforms the baseline method and two state-of-the-art methods and avoids the tedious procedure required for the sphere approximation.
}

\begin{table}[H]
\caption{\DIFaddFL{Results of baseline method, other state-of-the-art methods and our proposed approach on reduced test set of UP-3D % Please define if appropriate.
 }\cite{lassner2017unite}\DIFaddFL{. }}
\centering
%DIF > % \tablesize{} %% You can specify the fontsize here, e.g.,  \tablesize{\footnotesize}. If commented out \small will be used.
\begin{tabular}{ccc}
\toprule
\DIFaddFL{\textbf{Methods}	}& \DIFaddFL{\textbf{Per Vertex Error (mm)}     }& \DIFaddFL{\textbf{Percentage of Vertices in Self-Intersection}}\\
\midrule
  \DIFaddFL{RE only     }& \DIFaddFL{257.54		}& \DIFaddFL{15.22}\%	\\
  \DIFaddFL{RE+LR }\cite{wang2018pixel2mesh}   & \DIFaddFL{452.72		}& \DIFaddFL{7.64}\%	\\
  \DIFaddFL{RE+SA }\cite{pons2015metric}   & \DIFaddFL{186.98       }& \DIFaddFL{0.87}\%\\
  \DIFaddFL{RE+SPT (ours)    }& \DIFaddFL{\textbf{140.31}     }& \DIFaddFL{\textbf{0.23\%}}\\
\bottomrule  %Is the bold necessary? If not, please remove.
\end{tabular}
\label{tab:quantitative}
\end{table}

\DIFaddend \subsection{Analysis of Time Efficiency}
In order to evaluate the time efficiency of our method, we carried out experiments with a different number of detection rays and SMPL models. All experiments in this section were done on a laptop with Intel(R) Core(TM) i5-3230M processer. The most time-consuming part of our technique is self-intersection detection as is described in Algorithm \ref{alg:detect}. The highlight of our method is that the time complexity is linear to the number of triangles. The number of detection rays has almost no effect on the elapsed time as is shown in Table \ref{tab:rays}. As can be seen in the Table \ref{tab:rays}, the number of detection rays was increased from $128\times128$ to $2048\times2048$ but the elapsed time remained approximately constant.

\begin{table}[H]
\caption{Time consumed by an iteration with different number of detection rays. The number of SMPL models is 1.}
\centering
%% \tablesize{} %% You can specify the fontsize here, e.g.,  \tablesize{\footnotesize}. If commented out \small will be used.
\begin{tabular}{ccc}
\toprule
\textbf{Number of Detection Rays}	& \textbf{Elapsed Time (ms)}\\
\midrule
$128\times128$		& 53.84	\\
$256\times256$		& 54.96	\\
$512\times512$       & 56.76\\
$1024\times1024$     & 58.32\\
$2048\times2048$     & 59.28\\
\bottomrule
\end{tabular}
\label{tab:rays}
\end{table}

In the next experiment, we fixed the number of detection rays and changed the number of SMPL models to test the performance of our method with growing number of triangles. A single SMPL model has about 13,000 triangles. As is shown in Table \ref{tab:model}, the elapsed time grows about 30 ms for each additional SMPL model added to the mesh. Result of this experiment demonstrated that the time complexity of our proposed algorithm is linear to the number of triangles.

\begin{table}[H]
\caption{Time consumed by an iteration with different number of SMPL models. The number of detection rays is $512\times512$.}
\centering
%% \tablesize{} %% You can specify the fontsize here, e.g.,  \tablesize{\footnotesize}. If commented out \small will be used.
\begin{tabular}{ccc}
\toprule
\textbf{Number of SMPL Models}	& \textbf{Elapsed Time (ms)}\\
\midrule
1		& 56.76			\\
2		& 89.80			\\
3       & 119.25   \\
4     & 149.58  \\
5     & 186.76   \\
\bottomrule
\end{tabular}
\label{tab:model}
\end{table}

Our proposed method is obviously more computational compared with traditional methods, but it is worthwhile to apply this method because of the accuracy and generality of our approach. Moreover, the time efficiency of our method is totally acceptable according to the experimental results.

%%%%%%%%%%%%%%%%%%%%%%%%%%%%%%%%%%%%%%%%%%
\section{Conclusions}\label{sec:conclusion}

In this paper, we proposed a novel self-intersection penalty term for \DIFdelbegin \DIFdel{statical }\DIFdelend \DIFaddbegin \DIFadd{statistical }\DIFaddend body shape models to remove the self-intersection of the mesh by gradient-based optimization. Unlike most traditional approaches, our method does not require a hard-to-obtain differentiable penalty term, but~instead gradients are manually calculated. In the course of analysis, we have demonstrated that it is not necessary to derive differentiable expressions of a penalty term and gradients can be manually calculated from the perspective of geometry. Since no approximation is used in our method, self-intersection can be strictly removed. The highlight of our work is that our method applies to general meshes with different shapes and topology without the need to design a set of appropriate proxy geometries. Despite the fact that our proposed self-intersection penalty term is more time consuming than traditional approaches, the elapsed time of one iteration is totally acceptable according to the experimental results. The applications of our method are not limited to the \DIFdelbegin \DIFdel{statical }\DIFdelend \DIFaddbegin \DIFadd{statistical }\DIFaddend body shape models presented in this paper. Our proposed self-intersection penalty term can be incorporated into other 3D reconstruction problems based on a triangular mesh, such as the mesh-based 3D reconstruction described in~\cite{wang2018pixel2mesh,Kar_2015_CVPR,kato2018neural}.

\DIFdelbegin \DIFdel{An obvious limitation of our method is that there may be }\DIFdelend \DIFaddbegin \DIFadd{Our proposed approach has its limitations. When there are }\DIFaddend some triangles happened to be parallel to the detection rays\DIFdelbegin \DIFdel{. In this case}\DIFdelend , these triangles will not intersect with any detection rays no matter how dense the detection rays are. That is to say, vertices of triangles parallel to the detection rays may be mistakenly classified, and~further the gradients with respect to these vertices will be incorrect. \DIFaddbegin \DIFadd{We assume that each triangle and its vertices are located in the same region, this assumption does not apply to the situation where the mesh is sparse and in this situation undesirable consequences may be caused. }\DIFaddend Another limitation is that it is difficult to \DIFaddbegin \DIFadd{manually }\DIFaddend set the number of detection rays appropriately, such that all vertices in self-intersection can be detected and classified correctly \DIFaddbegin \DIFadd{with minimum memory consumption}\DIFaddend .

Future research directions of this work may include modifying the way detection rays are emitted to avoid incorrect results when there are triangles parallel to the detection rays. It may also include developing a strategy to set the number of detection rays appropriately and automatically. We are also interested in reducing the time complexity of our proposed method to make this approach more suitable for gradient-based optimization.

%%%%%%%%%%%%%%%%%%%%%%%%%%%%%%%%%%%%%%%%%%
\vspace{6pt}

%%%%%%%%%%%%%%%%%%%%%%%%%%%%%%%%%%%%%%%%%%
%% optional
%\supplementary{The following are available online at \linksupplementary{s1}, Figure S1: title, Table S1: title, Video S1: title.}


%%%%%%%%%%%%%%%%%%%%%%%%%%%%%%%%%%%%%%%%%%
\authorcontributions{Conceptualization, Z.W. and H.L.; Methodology, Z.W., H.L. and W.J.; Writing---original draft preparation, Z.W.; Writing---review and editing, Z.W., H.L. and W.J.; Visualization, Z.W. and H.L.; Supervision, W.J.; Funding acquisition, W.J., H.L. and L.C.}

%%%%%%%%%%%%%%%%%%%%%%%%%%%%%%%%%%%%%%%%%%
\funding{This research was funded by [the National Natural  Science Foundation of China] grant number [61633019], [the Public Projects of Zhejiang Province, China] grant number [LGF18F030002] and [Huawei innovation research program] grant number [HO2018085209].}

%%%%%%%%%%%%%%%%%%%%%%%%%%%%%%%%%%%%%%%%%%
%\acknowledgments{We would like to thank Mingxue Wang and Lin Cheng from 2012 Lab, Huawei Technologies for providing computation resources and technical support.}

%%%%%%%%%%%%%%%%%%%%%%%%%%%%%%%%%%%%%%%%%%
\conflictsofinterest{The authors declare no conflict of interest.}

\reftitle{References}
%\begin{thebibliography}{999}
% Reference 1
%\bibitem[Author1(year)]{ref-journal}
%Author1, T. The title of the cited article. {\em Journal Abbreviation} {\bf 2008}, {\em 10}, 142-149, doi:xxxxx.
% Reference 2
%\bibitem[Author2(year)]{ref-book}
%Author2, L. The title of the cited contribution. In {\em The Book Title}; Editor1, F., Editor2, A., Eds.; Publishing House: City, Country, 2007; pp. 32-58, ISBN.
%\end{thebibliography}

% The following MDPI journals use author-date citation: Arts, Econometrics, Economies, Genealogy, Humanities, IJFS, JRFM, Laws, Religions, Risks, Social Sciences. For those journals, please follow the formatting guidelines on http://www.mdpi.com/authors/references
% To cite two works by the same author: \citeauthor{ref-journal-1a} (\citeyear{ref-journal-1a}, \citeyear{ref-journal-1b}). This produces: Whittaker (1967, 1975)
% To cite two works by the same author with specific pages: \citeauthor{ref-journal-3a} (\citeyear{ref-journal-3a}, p. 328; \citeyear{ref-journal-3b}, p.475). This produces: Wong (1999, p. 328; 2000, p. 475)

%=====================================
% References, variant B: external bibliography
%=====================================
\externalbibliography{yes}
%\bibliography{refworks}

%%%%%%%%%%%%%%%%%%%%%%%%%%%%%%%%%%%%%%%%%%
%% optional
%\sampleavailability{Samples of the compounds ...... are available from the authors.}

%% for journal Sci
%\reviewreports{\\
%Reviewer 1 comments and authors’ response\\
%Reviewer 2 comments and authors’ response\\
%Reviewer 3 comments and authors’ response
%}

%%%%%%%%%%%%%%%%%%%%%%%%%%%%%%%%%%%%%%%%%%
\end{document}